\documentclass[runningheads]{llncs}
\pdfoutput=1
 
\usepackage{eccv}

\usepackage{eccvabbrv}

\usepackage{graphicx}
\usepackage{booktabs}

\usepackage{multirow}

\usepackage[accsupp]{axessibility}  

\usepackage{hyperref}

\usepackage{orcidlink}

\newcommand{\ours}{RingID}

\usepackage{amsmath}

\DeclareMathOperator*{\argmin}{arg\,min}

\usepackage{colortbl}
\definecolor{mygray}{gray}{.95}

\usepackage{tikz}
\usepackage{pgfplots}
\usepackage{caption}
\usepackage{bm}

\usepackage[misc]{ifsym}

\usepackage{ifthen}
\newboolean{camready}
\setboolean{camready}{false} 
\newcommand{\camreadyorarxiv}[2]{\ifthenelse{\boolean{camready}}{#1}{#2}}

\begin{document}

\title{\ours: Rethinking Tree-Ring Watermarking for Enhanced Multi-Key Identification} 

\titlerunning{RingID: Enhanced Multi-Key Identification}

\author{Hai Ci$^*$\orcidlink{0000-0001-7170-277X} \and
Pei Yang$^*$\orcidlink{0009-0000-3948-6915} \and
Yiren Song$^*$\orcidlink{0000-0002-7028-3347} \and
Mike Zheng Shou\textsuperscript{\Letter}\orcidlink{0000-0002-7681-2166}}

\authorrunning{H.~Ci et al.}

\institute{Show Lab, National University of Singapore\\
\email{cihai03@gmail.com} \quad \email{yangpei@u.nus.edu}\\ 
\email{yiren@nus.edu.sg} \quad \email{mike.zheng.shou@gmail.com}}

\maketitle

\def\thefootnote{*}\footnotetext{Equal Contribution. $\textsuperscript{\Letter}$Corresponding Author.}\def\thefootnote{\arabic{footnote}}

\begin{abstract}
  We revisit {\it Tree-Ring Watermarking}, a recent diffusion model watermarking method that demonstrates great robustness to various attacks. 
  We conduct an in-depth study on it and reveal that the distribution shift unintentionally introduced by the watermarking process, apart from watermark pattern matching, contributes to its exceptional robustness. Our investigation further exposes inherent flaws in its original design, particularly in its ability to identify multiple distinct keys, where distribution shift offers no assistance. 
  Based on these findings and analysis, we present {\it RingID} for enhanced multi-key identification. It consists of a novel multi-channel heterogeneous watermarking approach designed to seamlessly amalgamate distinctive advantages from diverse watermarks. Coupled with a series of suggested enhancements, {\it RingID} exhibits substantial advancements in multi-key identification. Github page: \url{https://github.com/showlab/RingID}

\end{abstract}
\section{Introduction}
\label{sec:introuction}
With the popularity of diffusion models, 
a vast number of high-quality diffusion generated images are circulating on the internet.
It has become increasingly important to identify AI-generated images to detect fabrication and to track and protect copyrights. Luckily, watermarking~\cite{an2024benchmarking,fernandez2023stable} offers a reliable solution.

Research on watermarking has a long history in computer vision~\cite{cox2007digital,steinberg2001identification}. A series of works~\cite{zhu2018hidden,fernandez2022watermarking,tancik2020stegastamp,fernandez2023stable,zhao2023recipe} imprint imperceptible watermarks on images through small pixel perturbations. 
Recently, \cite{zhao2023generative} demonstrates that pixel-level perturbations are provably removable by regeneration attacks. In contrast, {\it Tree-Ring Watermarking}~\cite{wen2023tree} has recently introduced a novel approach, proposing to imprint a specially designed tree-ring pattern into the initial noise of a diffusion model. Different patterns are also called different "keys". The verification of the watermark's existence involves recovering the initial diffusion noise from the image and comparing the extracted key with the injected one.
{\it Tree-Ring} is considered a semantic watermark since the watermark changes the layout of the generated image and semantically hidden in it. 
It has demonstrated strong robustness against regeneration attacks~\cite{zhao2023generative} and various other image transformations~\cite{an2024benchmarking,wen2023tree}, credited to the meticulously designed tree-ring pattern. 
However, previous research only studies its performance in the verification task, which aims at distinguishing between watermarked and unwatermarked images. It's still unclear if {\it Tree-Ring} can be used to distribute multiple distinct keys and distinguish between them, a.k.a. watermark identification, which is crucial for source tracing and user attribution. Will {\it Tree-Ring} still be strong and robust as in the verification task? In this work, we will explore and answer this question.

\begin{figure}[tb]
  \centering
  \includegraphics[width=\linewidth]{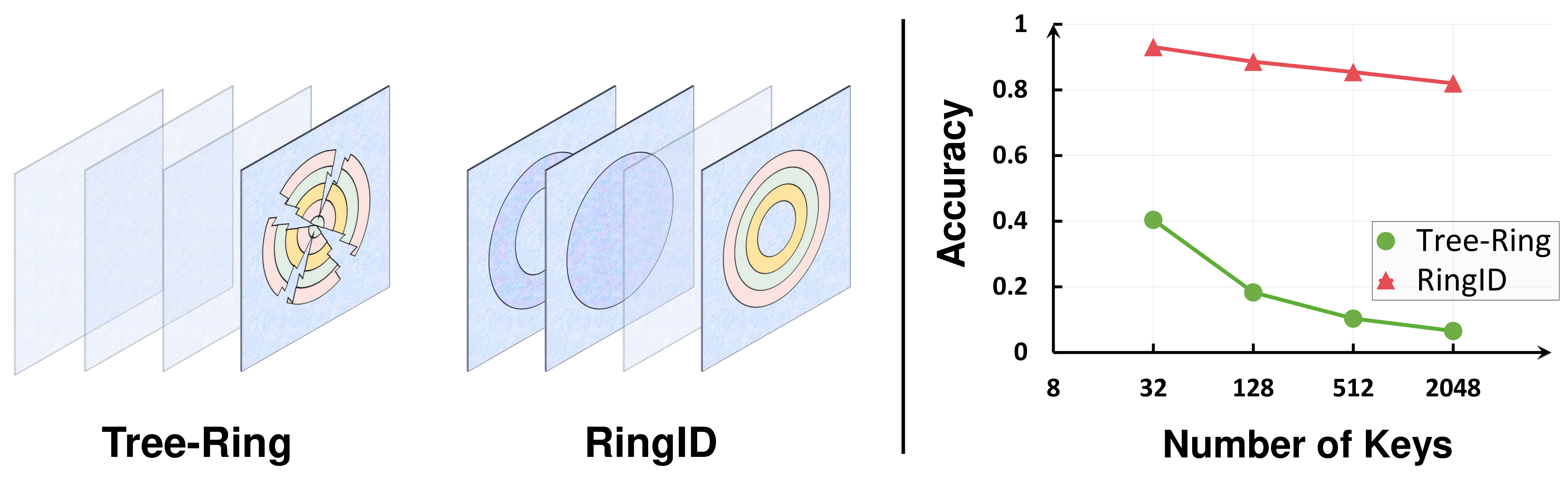}
  \caption{{\it Tree-Ring} imprints a broken watermark to a single channel. Our method {\it \ours} achieves stronger robustness by imprinting intact watermarks to multiple channels. In the task of multi-key identification, {\it \ours} shows overwhelming advantages.}
  \label{fig:teaser}
\end{figure}

We start by studying the source of {\it Tree-Ring}'s robustness under verification setting. For the first time, we uncover that the robustness of {\it Tree-Ring} stems not only from the carefully designed tree-ring pattern but also from the unintentionally introduced distribution shift during watermark imprinting. Such shift makes a big difference, especially in its robustness to Rotation and Crop/Scale image transformations, which cannot be handled by tree-ring's pattern design. More importantly, this overlooked power does not help in the identification scenario, because different watermarks experience the same shift.

We further evaluate {\it Tree-Ring} in the identification task and surprisingly find that {\it Tree-Ring} struggles to distinguish between even dozens of different keys. Identification accuracy is reported in~\cref{tab:identification}. The disappearance of the extra power of distribution shift has clearly exposed more design flaws in {\it Tree-Ring}. First, we find that tree-ring patterns are not discriminative enough to distinguish between different keys. This results in  {\it Tree-Ring}'s bad accuracy under various attacks or even under the clean setting. Meanwhile, the defective imprinting process leads to the injection of broken watermarks, further weakening its distinguishing ability. In addition, without the help of distribution shift, {\it Tree-Ring} is completely incapable of handling rotation and crop/scale transformations. Identification accuracy under these attacks is close to zero.

In order to address the above issues, we propose {\it \ours} for enhanced identification capability. It bases on a novel Multi-Channel Heterogeneous watermarking framework, which can amalgamate distinctive advantages from different types of watermarks to resist to various kinds of attacks. Coupled with discretization and lossless imprinting methods, {\it \ours} gains substantial improvement on distinguishing ability. What's more, for the specific attack such as rotation, we locate the flaws in {\it Tree-Ring}'s original design and propose systematic solutions, which improves the identification accuracy under rotation from nearly 0 to 0.86. 

To summarize our contributions:
\begin{enumerate}
  \item For the first time, we elucidate how the unintentional introduction of distribution shift substantially contributes to {\it Tree-Ring}'s robustness. We demonstrate this effect from both mathematical and empirical perspectives.
  \item We systematically explore {\it Tree-Ring}'s performance in the identification scenario and reveal its limitations through comprehensive analysis.
  \item We present {\it \ours}, a systematic solution that significantly enhances watermark verification and identification, delivering impressive results. The verification AUC improves from 0.975 to 0.995, while the identification accuracy rises from 0.07 to 0.82.
\end{enumerate}

\section{Related Work}
\label{sec:related_work}
\vspace{-0.2cm}

\subsection{Diffusion Models}
\vspace{-0.1cm}
Diffusion models ~\cite{ho2020denoising,song2020denoising,dhariwal2021diffusion} signify a swiftly advancing category of generative models. They define a forward diffusion process that gradually add small amount of Gaussian noise to the input data and learn to recover the clean data from the distortion~\cite{ho2020denoising}. Diffusion models initially demonstrate huge potential in generating high-resolution images~\cite{ho2020denoising,song2020score}, resulting in globally recognized products like Midjourney~\cite{Midjourney}, DALL-E~\cite{betker2023improving,DALLE2} and StableDiffusion~\cite{Rombach_2022_CVPR}. Thereafter, they are broadly applied to modeling various types of data, \eg video~\cite{wu2023tune,guo2023animatediff,singer2022make,blattmann2023stable,zhang2023show,song2024processpainter}, audio~\cite{kong2020diffwave}, 3d object~\cite{poole2022dreamfusion,liu2023zero}, humans~\cite{tevet2022human,ci2019optimizing,ci2023gfpose,zhu2023human}, etc.


\vspace{-0.1cm}
\subsection{Watermarking Diffusion Models}
When it comes to watermarking diffusion models, there are two different types of objectives: watermarking model weights~\cite{bansal2022certified,uchida2017embedding,zhang2018protecting,liu2023watermarking,zhao2023recipe} or generated content~\cite{yu2021artificial,zhu2018hidden,tancik2020stegastamp,zhang2019robust,fernandez2022watermarking,ci2024wmadapter}. 
Model weights watermarking methods aim to protect the intellectual property of the model itself and identify whether another model is an instance of plagiarism.  
While content watermarking methods protect the copyright of generated images~\cite{fernandez2023stable,yu2021artificial,yang2024steganalysis} or videos~\cite{zhang2019robust,luo2023dvmark} by adding imperceptible perturbations onto pixels.
They either imprint the watermark into the frequency via DFT~\cite{lin2001rotation}, DCT~\cite{cox2007digital,cui2023enhancing} and wavelet transformations~\cite{xia1998wavelet,cui2024estimation}, or directly onto pixels via an encoder network~\cite{zhu2018hidden,tancik2020stegastamp,guo2023practical,xiong2023flexible}. 
Recently, {\it Tree-Ring}~\cite{wen2023tree} proposes to imprint a tree-ring watermark pattern into the initial diffusion noise. 
It demonstrates strong robustness to various attacks. 
In this work, we delve into {\it Tree-Ring}, uncovering its source of power in watermark verification and pinpointing its vulnerability in identifying multiple keys. Subsequently, we introduce {\it \ours}, a systematic solution fortified with enhanced identification capability.

 


\section{Preliminaries}
\subsection{Threat Model}
{\bf Model Owner:} The owner of the diffusion model provides the image generation service through API access. He would like to imprint imperceptible watermarks into every generated image for copyright protection and source tracking. Given an arbitrary image, the model owner would like to check whether 
the image contains his own watermark
(a.k.a. {\it Watermark Verification}) and which watermark it is if he distributes multiple different watermarks (a.k.a. {\it Identification}). The goal of the model owner is to ensure high verification and identification accuracy regardless of any image transformations.

\noindent
{\bf Attacker:} The attacker uses the model owner's service to generate an image. He attempts to apply various image transformations such as rotation and JPEG compression to disrupt the watermark, in order to evade the detection and identification check of the model owner. Subsequently, he can claim ownership of the image and profit from it illegally

\subsection{Notation}
We denote the initial diffusion noise by $x_T$, and its frequency counterpart after FFT by $X_T$. $T$ is the diffusion step. Sometimes we may omit the subscript $T$ for simplicity. We denote the real part, imaginary part and conjugate symmetry of $X_T$ by $X_\text{re}$, $X_\text{im}$, and $X_\text{cs}$. In the context of StableDiffusion~\cite{Rombach_2022_CVPR}, $X_T \in R^{h \times w \times c}$ has 4 channels $\mathcal{C}=\{0, 1, 2, 3\}$. We denote the watermarked channels by $\mathcal{C}_w$, where $ \mathcal{C}_w \subset \mathcal{C}$.
We denote the imprinted key by $w$, the recovered key at detection time by $\hat{w}$, and the null watermark extracted from the unwatermarked image by $\hat{w}_\varnothing$. We use $\mathcal{W} = \{w_1, w_2, \ldots\}$ to denote a set of different keys. $\lvert\mathcal{W}\rvert $ is its capacity.

\vspace{-0.1cm}
\subsection{Task Formulation: Verification vs. Identification}
\vspace{-0.1cm}
In this section, we formulate the verification and identification task following the problem statement of \textit{Tree-Ring}~\cite{wen2023tree}.
The task of watermark verification is to detect the existence of watermarks. Verification algorithms attempt to distinguish between watermarked images carrying key $\hat{w}$ and unwatermarked images carrying null key $\hat{w}_\varnothing$. More formally, verification can be formulated as distinguishing between the distance to the reference key $w$: $d\left(\hat{w}, w\right) \neq d\left(\hat{w}_\varnothing, w\right)$. In {\it Tree-Ring}, $d$ is chosen to be $\ell_1$ distance and $w$ is the watermark to be imprinted.

When a set of different keys $w_i \in \mathcal{W}$ are distributed, the goal of watermark identification is to find the best match for the recovered key $\hat{w}$ among all candidates in $\mathcal{W}$, \ie $\text{ID}\left(\hat{w}\right) = \argmin_{i} \left[d\left( \hat{w}, w_i \right) \right]$. In practice, different users are usually assigned different keys. Thus, each generated image can be source tracked and attributed to a unique user via watermark identification.


\begin{figure}[tb]
  \centering
  \includegraphics[width=\linewidth]{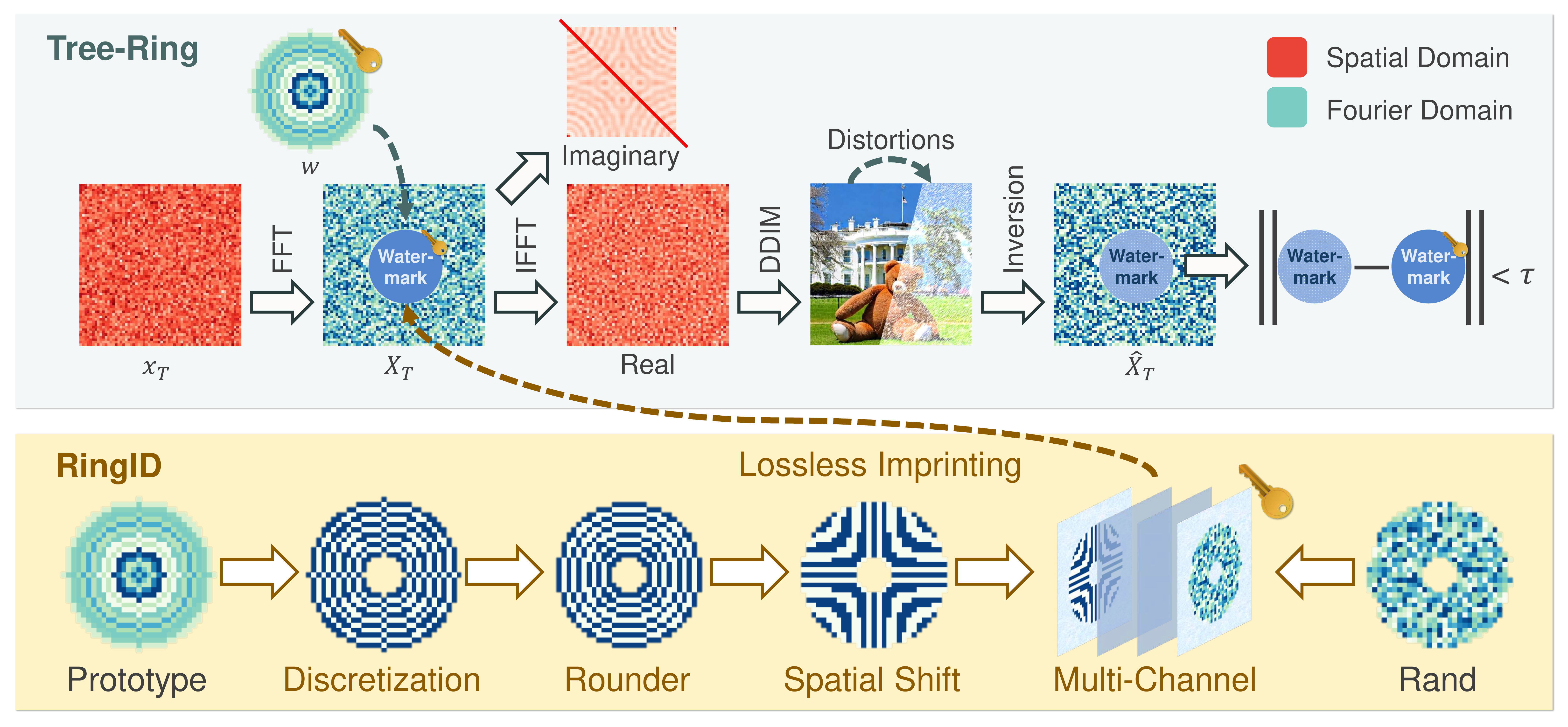}
  \caption{Framework of the watermarking process. {\it \ours} introduces a series of approaches that can help to imprint a lossless and robust watermark. In contrast, {\it Tree-Ring} injects a lossy and less robust watermark.}
  \label{fig:pipeline}
\end{figure}

\section{Delve into Tree-Ring Watermarking}
\label{sec:method_revisit}
In this section, we will reveal why {\it Tree-Ring}~\cite{wen2023tree} is robust to various image transformations, even some of them are not considered by design. 
We further evaluate {\it Tree-Ring} in the multi-key identification task and pinpoint its vulnerabilities.

\subsection{Recap of Tree-Ring}
{\it Tree-Ring}~\cite{wen2023tree} originally studies the verification setting. Its goal is to distinguish watermarked images from non-watermarked images. 
As illustrated in~\cref{fig:pipeline}, {\it Tree-Ring} transforms the initial diffusion noise $x_T$ into the frequency $X_T$ and injects a tree-ring pattern $w$ into the center of $X_T$, where the value of each ring in pattern $w$ is sampled from Gaussian. It then transforms the watermarked $X_T$ back to the spatial domain, retaining only the real part to feed the diffusion network for image generation. Notably, the accompanying imaginary part is discarded to be compatible with the network input. For detection of the watermark, {\it Tree-Ring} inverts the diffusion process to estimate $x_T$ from an image, transforms it to the frequency domain, and extracts the watermark pattern $\hat{w}$. 
Similarly, nonsense pattern $\hat{w}_\varnothing$ can be extracted from the non-watermarked image. 
{\it Tree-Ring} distinguishes between these two $\ell_1$-to-reference distances: $\lVert\hat{w} - w\rVert_1 $ and $ \lVert\hat{w}_\varnothing - w\rVert_1$ to determine whether an image is watermarked or not.

\subsection{Distribution Shift in Watermarking}
\label{sec:analysis}
{\it Tree-Ring}'s detection algorithm bases on the intuition that the recovered pattern $\hat{w}$ from a watermarked image should match the injected reference pattern $w$ to some extent, leading to a smaller $\ell_1$-to-reference distance, \ie $\lVert\hat{w} - w\rVert_1 < \lVert\hat{w}_\varnothing - w\rVert_1$.

However, we have discovered that apart from the pattern matching, another crucial factor contributes to the separation of $\lVert\hat{w} - w\rVert_1$ {\it v.s.} $\lVert\hat{w}_\varnothing - w\rVert_1$. This factor is the distribution shift introduced by discarding the imaginary part during the watermark injection process, which is highlighted in~\cref{fig:pipeline}. 
This affects all watermarked images, further reducing the expectation of $\mathbb{E}\left[\lVert\hat{w} - w\rVert_1\right]$. In other words, it shifts the distribution of $\lVert\hat{w} - w\rVert_1$ towards a smaller value, making it easier to distinguish between $\lVert\hat{w} - w\rVert_1$ and $\lVert\hat{w}_\varnothing - w\rVert_1$.

Mathematically, we prove that the shift factor is $\frac{\sqrt{3}}{2}$ in a general and simplified setting.
Let $\hat{w}_{n}$ denote the recovered watermark that has never been shifted, thus we have the following:

\begin{equation}
    \mathbb{E}\left[\lVert\hat{w} - w\rVert_1\right] = \frac{\sqrt{3}}{2}\mathbb{E}\left[\lVert\hat{w}_{n} - w\rVert_1\right].
\end{equation}

\begin{figure}[tb]
  \centering
  \includegraphics[width=0.9\linewidth]{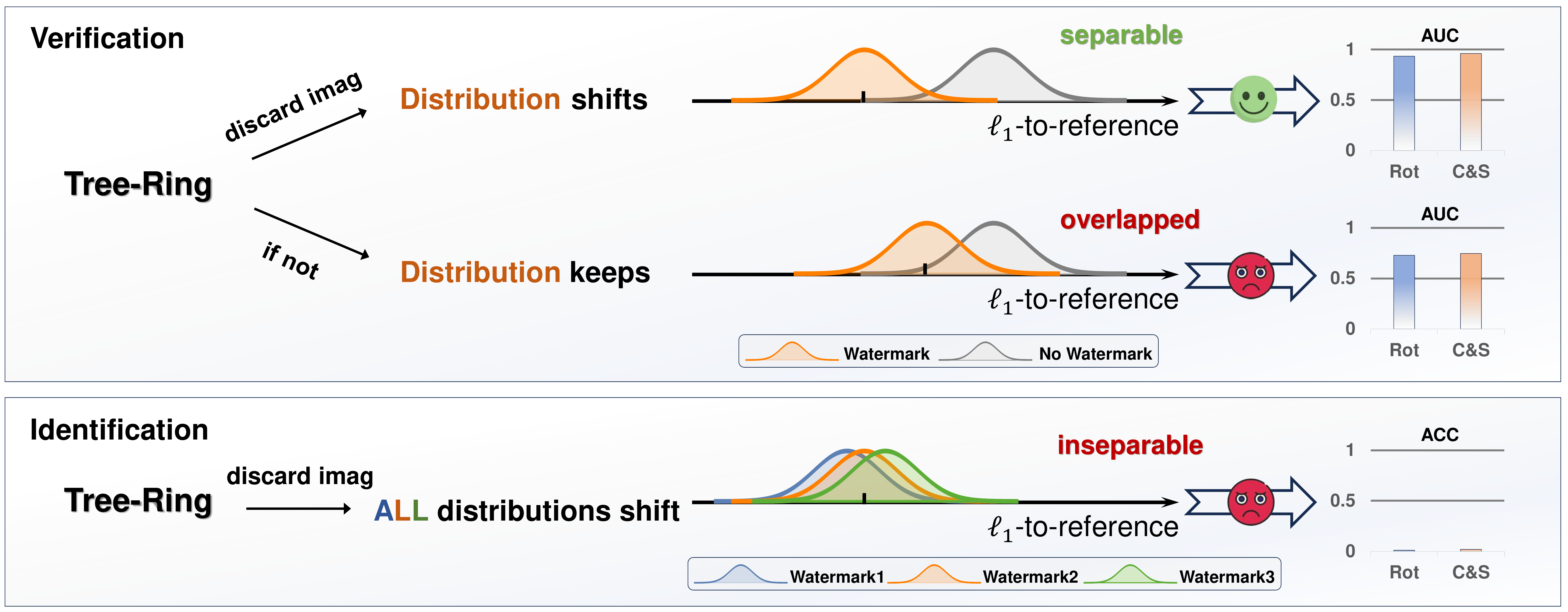}
  \caption{The effect of distribution shift in both verification and identification tasks. Discarding the imaginary part during watermark imprinting causes the distribution shift in $\ell_1$ distance. Distribution shift aids verification but not identification.}
  \label{fig:motivation}
\end{figure}

Detailed derivation is elaborated in\camreadyorarxiv{ Supplementary A.}{~\cref{sec:supp_shift}.}
Actually, we find any operation leading to a sufficient distribution shift can be considered a good stand-alone watermarking approach. We demonstrate the operation of discarding the imaginary part performs well on its own in\camreadyorarxiv{ Supplementary B.}{~\cref{sec:supp_standalone_watermark}.}

\subsubsection{Effect in Different Tasks and Attacks}
\cref{fig:motivation} illustrates the effect of distribution shift under different tasks. In \emph{Verification}, it helps distinguish between watermarked and non-watermarked images. Next, we go one step further and consider the situation when the image undergoes different attacks like JPEG or Resize. If the tree-ring pattern itself can cope with a certain attack, the improvement brought by the distribution shift will be small, because the two distributions are already separable even without the help of shift. On the contrary, if the tree-ring pattern itself cannot deal with a certain attack, the improvement brought by the distribution shift will be significant. Empirically (\camreadyorarxiv{Supplementary A.6}{\cref{subsec:supp_shift_in_real}}), we find distribution shift helps a lot under Rotation and Crop/Scale attacks. This implies that the original design of tree-ring cannot handle these two attacks. 
When it comes to \emph{Identification}, things are a little different. Since different watermarks experience the same distribution shift, discarding the imaginary part doesn't help distinguishing them.

This sparks our curiosity about {\it Tree-Ring}'s performance in identification, since distribution shift doesn't help and identification is more difficult. We would like to know if {\it Tree-Ring} still works under Rotation and Crop/Scale attacks. In the next section, we will explore these questions.

\subsection{Is Tree-Ring Good at Identification?}
\label{subsec:identification_analysis}
We report the performance of {\it Tree-Ring} when identifying different numbers of keys in~\cref{tab:identification}. {\it Tree-Ring} struggles to identify as few as 32 keys. What's more, identification accuracy is nearly zero under Rotation and Crop/Scale attacks.

We summarize 2 reasons as follows: (1) Distribution shift doesn't help. As tree-ring watermark cannot handle Rotation and Crop/Scale attacks, identification accuracy under these attacks drops to nearly zero. (2) Tee-ring pattern doesn't have enough distinguishing ability. Since identification is more difficult than verification, poor distinguishing ability results in poor performance under most attacks or even under no attack setting.



\subsection{Why is Tree-Ring Vulnerable to Rotation and Crop/Scale?}
All previous experimental results and analysis show one thing, tree-ring pattern cannot handle these two attacks. In this section, we will see what happens.

\noindent
{\bf Vulnerable to Crop \& Scale}
Scaling an image by a factor of $a$ corresponds to compressing the frequency spectrum $X\left[ u,v \right]$ in both magnitudes and positions: $\frac{1}{a^2}X\left[ \frac{u}{a}, \frac{v}{a} \right]$.
Cropping an image modulates the frequency magnitudes by a sinc function. It is straightforward that these transformations easily disrupt the tree-ring pattern on frequency domain. As shown in \cref{fig:method_details}(e), the pattern is completely disrupted. 

\noindent
{\bf Vulnerable to Rotation}
Although {\it Tree-Ring} is designed as ring patterns to seek rotation robustness, we find the existing design ignores many details, leading to no rotation robustness. In short, a combination of corner cropping when rotating, a lossy injection process, and the rotationally asymmetric shape cause the problem. We will explain in detail and propose solutions in~\cref{subsec:method_rotation}.

\section{\ours: Enhanced Watermarking for Identification}
\label{sec:method_ours}
In this section, we show how our proposed method {\it \ours} advances in identifying multiple distinct keys. {\it \ours} originates from {\it Tree-Ring} but incorporates systematic enhancements. We introduce a  multi-channel heterogeneous watermarking framework, a discretization approach and a method of lossless imprinting to generally improve its distinguishing ability. We also discuss a series of designs regarding stronger rotation robustness and increased capacity.

\subsection{Multi-Channel Heterogeneous Watermarking}
Multi-channel heterogeneous (MCH) watermarking serves as a general framework to comprehensively improve watermarking robustness to a battery of attacks. Intuitively, we imprint different types of watermarks on different channels of the initial noise to amalgamate their distinctive advantages, as illustrated in \cref{fig:teaser}. During matching, we calculate the minimum of channel-wise $\ell_1$ distance to the reference watermark for all watermarked channels:
\begin{equation}
\label{eq:heter_watermark}
    \text{ID}\left(\hat{w}\right) = \argmin_{i} \left \{ \min_{c \in C_w}\left[\lambda_c \lVert\hat{w}^c - w^c_i\rVert_1 \right] \right \}
\end{equation}
where $c$ denotes the channel index and traverses all watermarked channels $C_w$. $i$ denotes the index of candidate reference watermarks. $\lambda_c$ is a channel-wise normalizing factor. 

\cref{eq:heter_watermark} bases on the intuition that if a watermark is robust to a certain attack, then its distance from the ground-truth reference is smaller than that of other watermarks that are not robust.
Ideally, it enables adaptive selection of the most robust watermark under different attacks. 
Notably, the total capacity of MCH is the minimum of each watermark. Therefore, it's recommended to use a watermark pattern with sufficient capacity on each channel.
At the same time, in order to reduce the impact on the generation quality, we need to carefully select the type of watermarks for combination. We find the Gaussian noise pattern is a perfect candidate. It has infinite capacity and the same distribution as the initial noise. Meanwhile, it has good robustness to non-geometric attacks~\cite{wen2023tree} and is complementary to the tree-ring pattern. So {\it \ours} explores the combination of a Gaussian noise watermark and a tree-ring watermark.
Empirical study in \cref{subsec:ab_robust} demonstrates that this combination is able to perfectly amalgamate unique advantages from the noise and ring watermark.

\begin{figure}[tb]
  \centering
  \includegraphics[width=.9\linewidth]{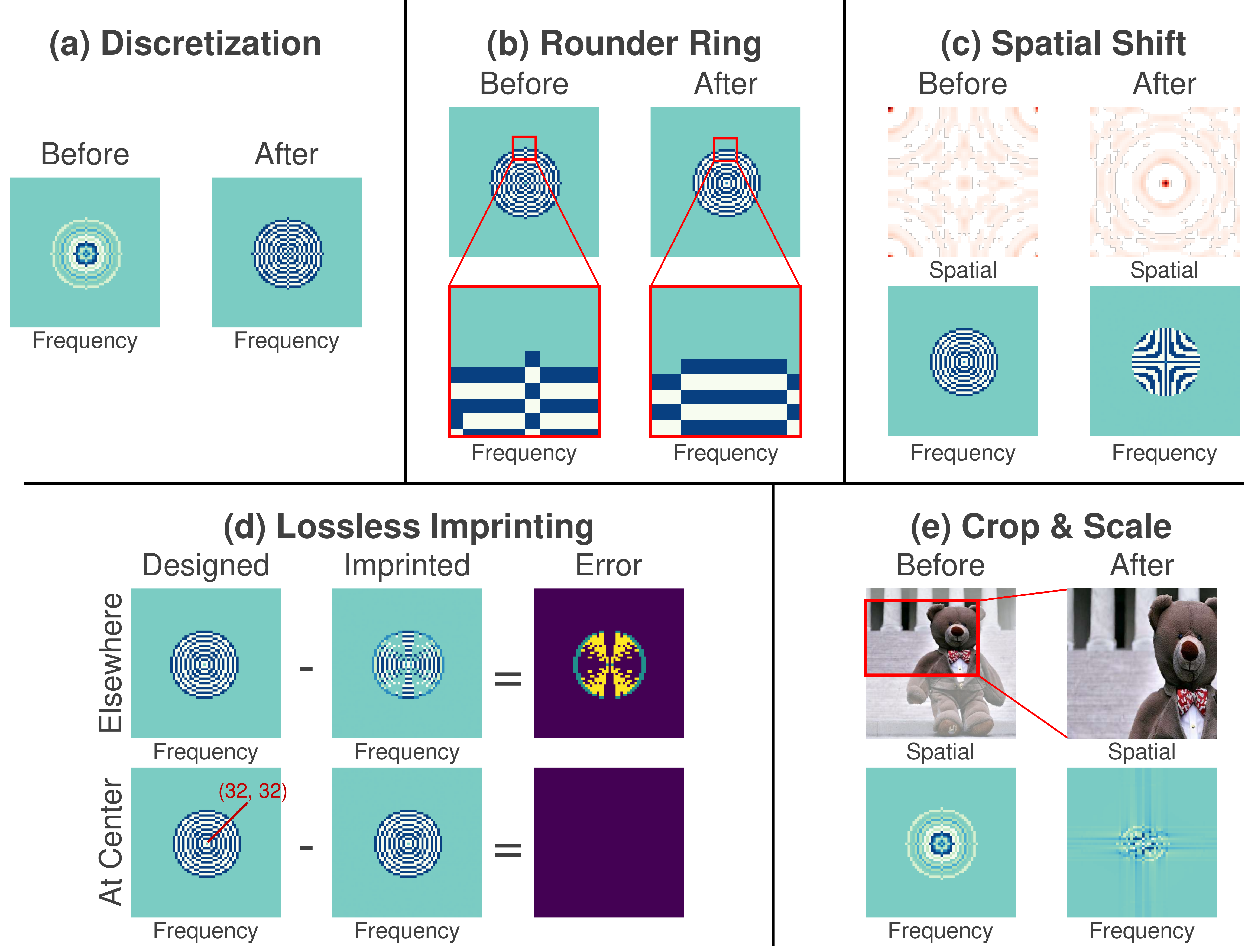}
  \caption{Detailed illustration of different acts and their impacts. (a) \textbf{Discretization}: Discretize values to $\pm \alpha$ for each ring, increasing the distinguishing ability of the watermark. (b) \textbf{Rounder ring:} Render smoother edges and reduce artifacts. Increase rotational robustness. (c) \textbf{Spatial Shift:} Shift energy on the corners to the center, avoiding potential cropping during rotation. (d) \textbf{Lossless Imprinting:} Ensure the imprinted pattern the same as the original design. (e) \textbf{Crop \& Scale:} Zooming in the spatial domain compresses the imprinted pattern in the frequency domain, causing pattern mismatch and hinders identification.} 
  \label{fig:method_details}
\end{figure}

\subsection{Being Invariant to Rotation}
\label{subsec:method_rotation}
Although the tree-ring pattern is designed to be rotationally symmetric, it exhibits unexpected vulnerability to rotation as shown in \cref{tab:identification}. 
Below, we identify the causes of the issue and propose a series of measures to address it.

\subsubsection{Spatial Shift to Avoid Corner Cropping} 
We visualize the imprinted tree-ring watermark in spatial domain in \cref{fig:method_details}(c). Due to the properties of FFT, tree-ring patterns in the frequency tends to exhibit concentrated energy in the four corners in the spatial domain, rendering them vulnerable to cropping during rotation. 
We propose to circularly shift the watermark pattern in the spatial domain to the center of the image, thus the majority of watermark information is kept while rotating. For an $N\times N$ latent image, we circularly shift the pattern by $\frac{N}{2}$ pixels in both width and height dimension. This is equivalent to directly multiplying the frequency watermark with a chessboard pattern given by $H[u,v]=e^{j\frac{\pi}{2}(u+v)}=(-1)^{u+v}$. The change of the watermark can be clearly observed in \cref{fig:method_details}(c). 
Although spatial-shifting helps a lot for rotation invariance, we noticed that this often leads to the generation of circular artifacts at the center of the image. Keeping generation quality in mind, we further multiply the shifted pattern by a factor $\eta$ to suppress the peak at the center. $\eta$ trades-off the robustness, as smaller $\eta$ weakens the watermark pattern. In practice, we find $\eta \in [0.8, 0.9]$ strikes a good balance between image quality and robustness.

\subsubsection{Lossless Imprinting}
The operation of discarding the imaginary part during watermarking not only shifts the distribution of $\ell_1$-to-reference, but also disrupt the pattern, resulting in the loss of its rotational symmetry. \cref{fig:method_details}(d) visualizes the actually carried pattern.
More concretely, recall from \cref{fig:pipeline}, when the  watermarked noise $X[u,v]$ is transformed back to the spatial $x[m,n]$, only its real part $x_\text{re}[m,n]$ is kept for image generation. From a frequency domain perspective, this is equivalent to generation with its conjugate symmetry $X_\text{cs}[u,v]$, which is not necessarily equal to $X[u,v]$.
To leverage the rotational symmetry of the ring patterns, we attempt to eliminate the information loss during imprinting by solving $X[u,v]=X_\text{cs}[u,v]$. The results indicate that $X_\text{re}[u,v]$ should be an even function and $X_\text{im}[u,v]$ should be an odd function about the Fourier center. To satisfy these constraints, we imprint the tree-ring pattern only in the real part and leave the imaginary part blank. We also align the center of the tree-ring pattern with the Fourier center $(32, 32)$. Originally, the tree-ring pattern was put at $(31, 32)$ and imprinted on both the real and the imaginary.

\subsubsection{Making A Rounder Ring}
Rings are originally determined by locating pixels equidistant from a give circle center $\left(x_0, y_0\right)$: $\left(x-x_0\right)^2 + \left(y-y_0\right)^2 = r^2$. We observed that this simple approach does not yield a well-rounded circular ring, particularly in images with lower resolutions like initial noise of shape $64 \times 64$. Artifacts and alias are shown in \cref{fig:method_details}(b). Although it is centrosymmetric, it lacks rotational symmetry. 
Well, the solution is quite straightforward: drawing a rounder ring.
There are a number of tools that can help draw a much rounder ring. Here, we employ a very simple idea to get a rounder ring. 
We place a white pixel on a black background at a distance $r$ from the rotation center. By rotating the low resolution image 360 degrees and recording the trajectory of this pixel, we obtain a rounder ring with a radius of $r$. Despite its simplicity, empirical results in~\cref{tab:Ablation—rob} show that it is crucial for rotational invariance.

\subsection{Discretization for Enhanced Distinguishability}
{\it Tree-Ring} samples values for each ring from a Gaussian distribution to keep consistency with the distribution of initial noise. We find that this random sampling makes it exceedingly challenging to distinguish between two keys. To address this issue, we propose a discretization approach, specifying that the value for each ring can only be either $\alpha$ or $-\alpha$, illustrated in \cref{fig:method_details}(a). Consequently, the watermark pattern for $n$ rings has a capacity of $2^n$. While this seemingly reduces the upper limit of watermark capacity, it significantly enhances its valid capacity, ensuring optimal utilization of all available slots. Our experiments reveal that setting $\alpha$ as the standard deviation of the initial noise ensures substantial distinguishablity between different keys while reducing the impact on generation quality to a small level.

\subsection{Increasing Capacity}
We consider two simple approaches to increase capacity: adjusting the number of rings in a single channel or imprinting rings in extra channels. We evaluate the advantages and drawbacks of both methods in \cref{subsec:discussion}. In summary, increasing the number of rings $|R|$ in a single channel can effectively boost capacity, leading to $2^{|R|}$ keys. However, excessive ring count results in a noticeable decline in robustness and generation quality. Imprinting watermarks in multiple channels exponentially increases capacity by $n^{|c|}$, where $c$ is number of channels and $n$ is the capacity of a single channel. However, it's likely to introduce ring-like artifacts in the center of generated images. Therefore, we opt to place the watermark in a single channel and enhance capacity by adjusting the range of rings.

\section{Experiment}
\label{sec:experiment}
\subsection{Experimental Setup}
Following {\it Tree-Ring}~\cite{wen2023tree}, we experiment with the state-of-the-art opensource diffusion model StableDiffusion-V2~\cite{Rombach_2022_CVPR}. By default, {\it \ours} imprints a tree-ring watermark with radius 3-14 on the channel 3 of the initial noise and a Gaussian noise watermark on channels 0. We adopt discretization (to 64 and -64), spatial shift, rounder drawing, and lossless imprinting as ordered in~\cref{fig:pipeline}. We report the ROC-AUC between 1000 watermarked and 1000 unwatermarked images as the verification metric. In identification, we report identification accuracy with different numbers of keys. 
Evaluation is performed under various image distortions, following the original setup of {\it Tree-Ring}~\cite{wen2023tree}, \ie, JPEG 25, rotation $75^{\circ}$, $75\%$ random C\&S, blur kernel 8, noise std 0.1, brightness [0, 6]. 
We assess the generation quality with CLIP scores~\cite{radford2021learning} by OpenCLIP-ViT/G~\cite{cherti2023reproducible} between the generated image and the prompts to measure text-image alignment, and Frechet Inception Distance (FID)~\cite{heusel2017gans} to quantify the similarity between generated and natural images. 
we generate 5000 images and evaluate FID on the COCO 2017 training dataset~\cite{lin2014microsoft}. We set both generation and inversion steps to 50.

\subsection{Comparison with Baselines}
\noindent
\textbf{Robustness}
In verification (\cref{tab:verification}), both {\it \ours} and {\it Tree-Ring} perfectly distinguish between watermarked and unwatermarked images when no attacks are performed, achieving AUC score of 1. However, {\it RingID} exhibit stronger robustness to various image distortions than {\it Tree-Ring}, achieving near-perfect average AUC score, even though it is not optimized for verification. 
In the more challenging identification task, (\cref{tab:identification}), {\it \ours} significantly outperforms {\it Tree-Ring}. When identifying 32 keys, {\it Tree-Ring}'s average identification accuracy (excluding C\&S) is 0.465, while {\it \ours}'s accuracy reaches up to 0.992. As the number of keys increases to 2048, this gap widens drastically (0.077 {\it v.s.} 0.942), further highlighting {\it \ours}'s superior robustness and larger valid key capacity.

Notably, both methods struggle with the C\&S distortion in identification (\cref{tab:identification}). This aligns with our analysis in \cref{sec:analysis}. Scaling in the spatial domain causes inverse scaling in the frequency domain, leading to pattern disruption and rendering identification challenging under such distortions. However, in verification, with the help of distribution shift,
watermarked images can still be effectively distinguished from unwatermarked ones under C\&S. Both {\it Tree-Ring} and {\it \ours} achieve average AUC higher than 0.97.

\noindent
\textbf{Image Quality}
{\it \ours} achieves strong performance without compromising the generation quality. It maintains comparable text-image alignments with {\it Tree-Ring}. Both methods achieve similar CLIP scores (0.364 {\it v.s.} 0.365). {\it \ours} obtains an FID of 26.13, which is also comparable to Tree-Ring's FID of 25.93.

\begin{table}[t]
    \centering
        \caption{Comparison with {\it Tree-Ring} in the verification task. The table shows the ROC-AUC values under various image distortions and CLIP Scores.}
        \label{tab:verification}
        \setlength{\tabcolsep}{2pt}
        \resizebox{\linewidth}{!}{%
        \begin{tabular}{c|c|ccccccc|c|c} 
        \toprule[1.1pt]
        \textbf{Methods} & \textbf{Ring Radius} & \textbf{Clean} & \textbf{Rotate} & \textbf{JPEG} & \textbf{C\&S} & \textbf{Blur} & \textbf{Noise} & \textbf{Brightness} & \textbf{Avg} & \textbf{CLIP Score}\\
        \midrule[0.75pt]
        \multirow{1}{*}{{\it Tree-Ring}}
        &0-10 &1.000 &0.935 &0.999 &0.961 &0.999 & 0.944 & 0.983 &0.975 & 0.364 \\
        \midrule[0.75pt]
        \multirow{2}{*}{{\it \ours}}
           &0-10 &1.000 &1.000 &1.000 & 0.979 &0.994 &0.969 &0.991 & 0.990 &0.359\\
           &3-14 &1.000 &1.000 &1.000 & 0.987 &0.989 &0.998 &0.994 & \textbf{0.995} & \textbf{0.365}\\
        \bottomrule[1.1pt]
        \end{tabular}}
    \vspace{-0.3cm}
\end{table}


\begin{table}[tb]
  \caption{Comparison with {\it Tree-Ring} in the identification task. We report the identification accuracy under various distortions for different numbers of keys.}
  \label{tab:identification}
  \centering
  \setlength{\tabcolsep}{2pt}
  \resizebox{\linewidth}{!}{%
  \begin{tabular}{c|c|ccccccc|cc} 
    \toprule[1.1pt]
    \textbf{Methods} & \textbf{\#Keys} & \textbf{Clean} & \textbf{Rotate} & \textbf{JPEG} & \textbf{C\&S} & \textbf{Blur} & \textbf{Noise} & \textbf{Brightness} & \textbf{Avg} & \textbf{\texorpdfstring{Avg\textsubscript{noC\&S}}{Avg no C\&S}} \\
    \midrule
    \multirow{3}{*}{{\it Tree-Ring}} & 32 & 0.790 & 0.020 & 0.420 & 0.040 & 0.610 & 0.530 & 0.420 & 0.404 & 0.465 \\
    & 128 & 0.450 & 0.010 & 0.120 & 0.020 & 0.280 & 0.230 & 0.170 & 0.183 & 0.210  \\
    & 2048 & 0.200 & 0.000 & 0.040 & 0.000 & 0.090 & 0.070 & 0.060 & 0.066 & 0.077  \\
    \midrule
    \multirow{3}{*}{{\it \ours}} & 32 & 1.000 & 1.000 & 1.000 & 0.530 & 0.990 & 1.000 & 0.960 & \textbf{0.926} & \textbf{0.992}  \\
    & 128 & 1.000 & 0.980 & 1.000 & 0.280 & 0.980 & 1.000 & 0.940 & \textbf{0.883} & \textbf{0.983}  \\
    & 2048 & 1.000 & 0.860 & 1.000 & 0.080 & 0.970 & 0.950 & 0.870 & \textbf{0.819} & \textbf{0.942}  \\
    \bottomrule[1.1pt]
  \end{tabular}}
\end{table}

\subsection{Ablation Study}
\label{subsec:ab_robust}
\noindent
\textbf{Components of RingID}
\cref{tab:Ablation—rob} ablates all proposed components of {\it \ours}. We can find that all designs are crucial for the final success of {\it \ours}. Spatial shift, lossless imprinting, and discretization play a significant role in the robustness of rotation. While the use of rounder rings also increase the accuracy by approximately 40\% (from 0.620 to 0.860). Notably, spatial shift sacrifices the robustness to blur and brightness attacks for rotation robustness.

\noindent
\textbf{Heterogeneous Watermarking}
{\it \ours} employs the combination of a Gaussian noise watermark and a ring watermark. Here, we ablate the choice of imprinting the noise watermark on different channels. By comparing the first 3 rows of 
\cref{tab:heterogeous}, we find that heterogeneous watermarking can effectively combine the advantage from both the noise watermark and the ring watermark. 
Among single-channel configurations, imprinting noise watermark on channel 0 (0.819) achieves the best performance. The combination of channels 0 and 1 yields the highest accuracy among all trials. Interestingly, imprinting noise watermarks on all available channels (0, 1, 2) decrease the accuracy from 0.830 to 0.807.


\begin{table}[tb]
  \caption{Ablation study on the effect of each proposed module in {\it \ours}. We report the identification accuracy for 2048 keys when removing individual components.}
  \label{tab:Ablation—rob}
  \centering
  \setlength{\tabcolsep}{3pt}
  \resizebox{\linewidth}{!}{%
  \begin{tabular}{l|ccccccc|c} 
    \toprule[1.1pt]
    \textbf{Settings} & \textbf{Clean} & \textbf{Rotate} & \textbf{JPEG} & \textbf{C\&S} & \textbf{Blur} & \textbf{Noise} & \textbf{Brightness} & \textbf{Avg} \\
    \midrule
    {\it \ours}               &1.000 &0.860 &1.000 &0.080 &0.970 &0.950 &0.870 &\textbf{0.819} \\
    - Spatial Shift      &1.000 &0.000 &1.000 &0.050 &0.990 &0.930 &0.940 &0.701 \\
    - Lossless Imprinting      &1.000 &0.010 &0.970 &0.160 &0.950 &0.980 &0.830 &0.700 \\
    - Rounder Ring    &1.000 &0.620 &0.990 &0.100 &0.890 &0.970 &0.850 &0.774 \\
    - Discretization   &0.980 &0.120 &0.380 &0.020 &0.450 &0.650 &0.390 &0.427 \\
    - Heterogeneous &1.000 &0.820 &0.940 &0.030 &0.960 &0.710 &0.720 &0.740 \\
  \bottomrule[1.1pt]
  \end{tabular}}
\end{table}

\begin{table}[tb]
  \caption{Ablation study on multi-channel heterogeneous watermarking. This table reports the identification accuracy when the noise watermark is imprinted on different channels (\textit{abbr.} CHs). ``-'' means the watermark is not imprinted.} 
  \label{tab:heterogeous}
  \centering
  \setlength{\tabcolsep}{4pt}
  \resizebox{\linewidth}{!}{%
  \begin{tabular}{cc|ccccccc|c} 
    \toprule[1.1pt]
    \textbf{Ring CHs} & \textbf{Noise CHs} & \textbf{Clean} & \textbf{Rotate} & \textbf{JPEG} & \textbf{C\&S} & \textbf{Blur} & \textbf{Noise} & \textbf{Brightness} & \textbf{Avg}\\
    \midrule
    -      &0           &1.000 &0.000 &1.000 &0.170 &0.990 &0.950 &0.920 &0.719 \\
    3      &-                 &1.000 &0.820 &0.940 &0.030 &0.960 &0.710 &0.720 &0.740 \\
    \midrule
    3      &0               &1.000 &0.860 &1.000 &0.080 &0.970 &0.950 &0.870 &\underline{0.819} \\
    3      &1               &1.000 &0.820 &1.000 &0.130 &1.000 &0.880 &0.770 &0.800 \\
    3      &2               &1.000 &0.840 &0.980 &0.100 &0.920 &0.990 &0.790 &0.803 \\
    \midrule
    3      &0, 1          &1.000 &0.850 &0.950 &0.150 &0.990 &0.990 &0.880 &\textbf{0.830} \\
    3      &0, 2          &1.000 &0.840 &0.990 &0.120 &0.950 &0.980 &0.870 &0.821 \\
    3      &0, 1, 2     &1.000 &0.830 &0.960 &0.160 &0.930 &0.980 &0.790 &0.807 \\
  \bottomrule[1.1pt]
  \end{tabular}}
\end{table}


\begin{figure}[ht]
    \begin{minipage}[t]{0.47\textwidth}
          \captionof{table}{Distributing 12 rings across different number of channels (\#CHs). This table reports the average identification accuracy and CLIP Score.}
          \label{tab:distribute_rings_to_multichannel}
          \vspace{0.1cm}
          \centering
          \setlength{\tabcolsep}{1pt}
          \resizebox{\linewidth}{!}{%
          \begin{tabular}{cc|cccc} 
            \toprule[1.1pt]
            \textbf{Ring Radius} &\textbf{\#CHs} & & \textbf{Avg} & & \textbf{CLIP Score}\\
            \midrule
            3-15 &1 & & \textbf{0.740}  & & \textbf{0.357} \\
            3-9  &2 & &0.737  & &0.351 \\
            3-7  &3 & &0.664  & &0.351 \\
            3-6  &4 & &0.544  & &0.340 \\
          \bottomrule[1.1pt]
          \end{tabular}}
    \end{minipage}
    \hfill
    \begin{minipage}[t]{0.45\textwidth}
            \strut\vspace*{-\baselineskip}\newline
            \centering
            \vspace{0.2cm}
            \includegraphics[width=\linewidth]{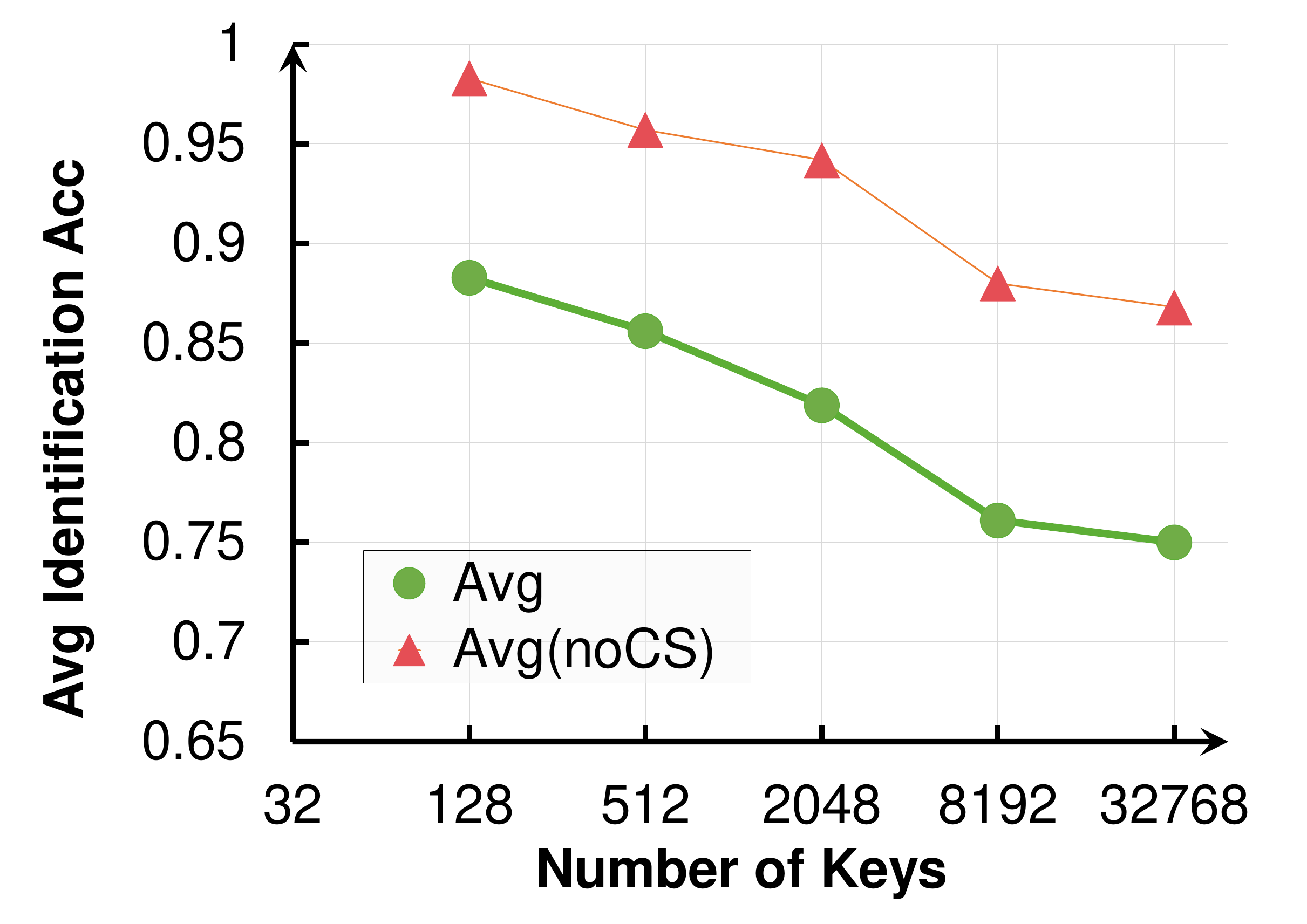}
            \vspace{-0.8cm}
            \captionof{figure}{Average identification accuracy {\it v.s.} number of keys.}
            \label{fig:example_caption}
    \end{minipage}
\end{figure}

\begin{table}[htb]
  \caption{Influence of ring radius on identification accuracy and text-image alignment.}
  \label{tab:betterring}
  \centering
  \setlength{\tabcolsep}{1pt}
  \resizebox{\linewidth}{!}{%
  \begin{tabular}{cccc|cccc|cccc} 
    \toprule[1.1pt]
    \textbf{Radius} & \textbf{\#Keys} & \textbf{Acc} & \textbf{CLIP Score} & \textbf{Radius} & \textbf{\#Keys} & \textbf{Acc} & \textbf{CLIP Score} & \textbf{Radius} & \textbf{\#Keys} & \textbf{Acc} & \textbf{CLIP Score} \\
    \midrule
    1-6   & 32   & 0.886 & 0.363 & 1-9   & 256  & 0.819 & 0.362 & 1-12  & 2048 & 0.786 & 0.361 \\
    2-7   & 32   & 0.876 & 0.366 & 2-10  & 256  & \textbf{0.856} & 0.358 & 2-13  & 2048 & 0.804 & 0.361 \\
    3-8   & 32   & \textbf{0.897} & 0.365 & 3-11  & 256  & 0.850 & 0.362 & 3-14  & 2048 & \textbf{0.819} & \textbf{0.368} \\
    4-9   & 32   & \textbf{0.897} & \textbf{0.368} & 4-12  & 256  & 0.847 & \textbf{0.363} & 4-15  & 2048 & 0.813 & 0.363 \\
    5-10  & 32   & 0.881 & 0.361 & 5-13  & 256  & 0.839 & 0.360 & 5-16  & 2048 & 0.803 & 0.360 \\
    \bottomrule[1.1pt]
  \end{tabular}}
\end{table}

\subsection{Discussion}
\label{subsec:discussion}
\noindent
\textbf{Where is the Best Place to Put Rings?}
We could imprint all rings on one channel or distribute them evenly across multiple channels. 
In~\cref{tab:distribute_rings_to_multichannel}, we fix the total number of rings to 12 and distribute them to 1-4 channels. we can find that performance drops as we distribute rings to more channels. What's more, imprinting rings on multiple channels cause artifacts to appear more frequently (shown in\camreadyorarxiv{ Supplementary C}{~\cref{sec:supp_failure_case}}). So given fixed capacity, it's better to imprint all rings on a single channel.
This raises another question: what are the optimal inner and outer radii if all rings are placed on a single channel? From~\cref{tab:betterring} we can see that  3-14 forms a good choice for 11 rings with capacity 2048. The optimal ring radius also varies for different ring numbers.

\noindent
\textbf{More Keys}
We would like to explore whether {\it \ours} can handle a larger number of keys. \cref{fig:example_caption} shows the average identification accuracy under various attacks {\it v.s.} number of keys. We observe that {\it \ours} maintains good accuracy even with a large number of keys. And identification accuracy linearly decreases as the number of keys exponentially increase.

\subsection{Qualitative Results}
\begin{figure}[tb]
  \centering
  \includegraphics[width=0.95\linewidth]{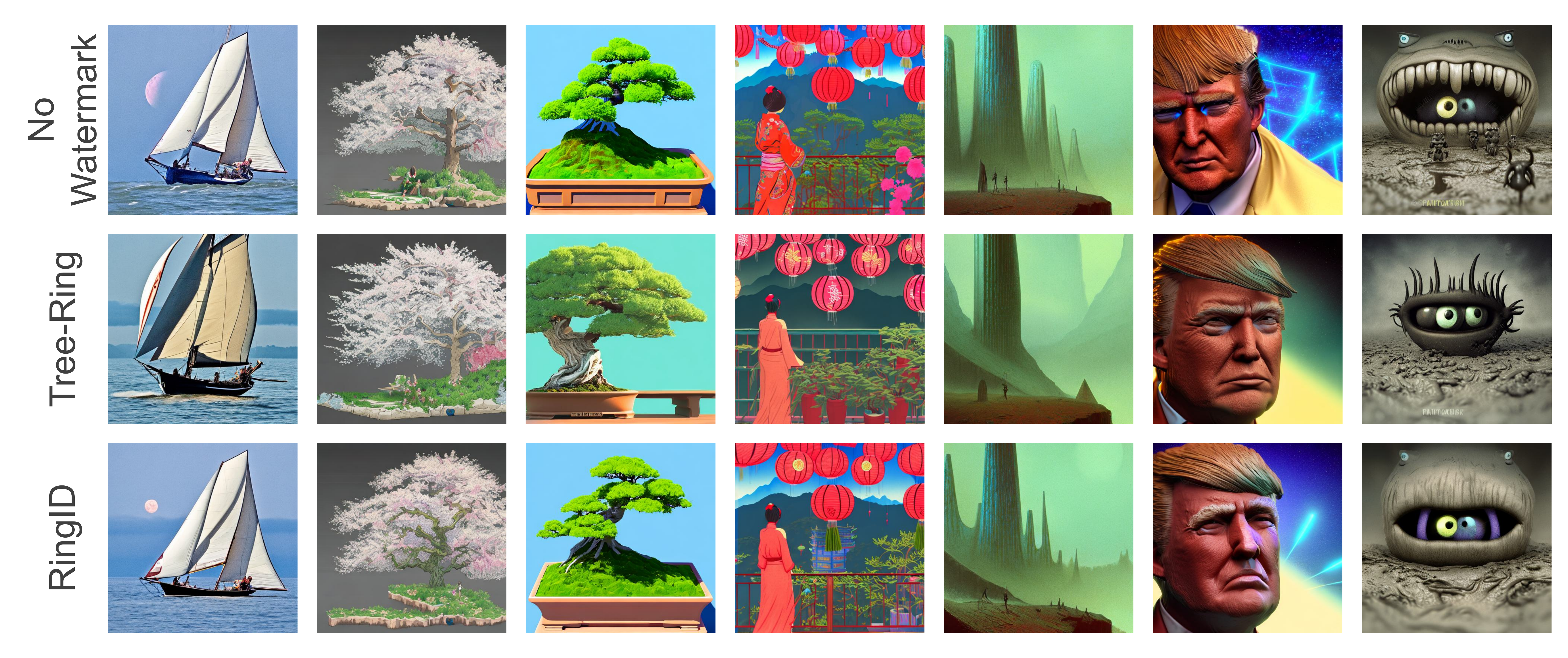}
  \caption{Qualitative results. The subtle changes in image layouts show that both {\it Tree-Ring} and {\it \ours} embed watermarks into the image's semantic content.}
  \label{fig:qualitative}
\end{figure}

As shown in \cref{fig:qualitative}, both {\it \ours} and {\it Tree-Ring} watermarking can generate images of good visual quality. Unlike low-level watermarking techniques, both {\it \ours} and {\it Tree-Ring} cause subtle changes in the image layout. However, {\it \ours} generates images with layout that is closer to the unwatermarked image. We owe this to avoiding watermarking on the low-frequency part (radius 0-3).

\section{Conclusion and Limitation}
\label{sec:conclusion}
We conduct a comprehensive investigation on {\it Tree-Ring Watermarking}, revealing its limitations in identifying multiple keys. We then propose {\it \ours} to enhance identification capability.

\noindent
\textbf{Limitation:} Under multi-key identification, one limitation is that both {\it Tree-Ring} and our {\it \ours} are vulnerable to cropping and scaling attacks. Future work may explore other transform domains to resist these attacks.


\noindent
\textbf{Acknowledgment:} This project is supported by the Ministry of Education, Singapore, under the Academic Research Fund Tier 1 (FY2022) Award 22-5406-A0001.



%
%
\bibliographystyle{splncs04}
\bibliography{main}

\clearpage

\appendix

\renewcommand\thesection{\Alph{section}}
\renewcommand\thesubsection{\thesection.\arabic{subsection}}

\section{Distribution Shift in Watermarking}
\label{sec:supp_shift}
The operation of discarding imaginary part during the watermarking process results in distribution shift in $\ell_1$-to-reference distance for any watermarked noise. In the following, we first demonstrate this from a mathematical view. In order to present this conclusion more clearly and concisely, we consider a more general case where all watermark pixels are i.i.d. sampled from the same watermark distribution as {\it Tree-Ring} but without a specific pattern for math simplicity. Then we study the more complicated real scenario by empirical experiments.


\subsection{Preliminaries}

We following the notation convention of \cite{dspbook, DIPBook}, representing 2D spatial domain signals using lower case letters indexed by $m, n$ (\textit{e.g.} $x[m, n]$) and 2D frequency domain signals using upper case letters indexed by $u, v$ (\textit{e.g.} $X[u, v]$). We use $\mathcal{F}$ and $\mathcal{F}^{-1}$ to denote DFT and inverse DFT, respectively. The energy of a signal $X[u, v]$ is defined as

\begin{equation}
    \mathcal{E}_X = \sum_{u,v} \lvert X[u,v]\rvert^2.
\end{equation}

A signal can be represented as the sum of a real-valued signal and a complex-valued signal, $X[u, v]=X_\mathfrak{re}[u, v]+jX_\mathfrak{im}[u, v]$. It can also be represented as the sum of a conjugate symmetric signal and a conjugate asymmetric signal, $X[u, v]=X_\text{cs}[u, v]+X_\text{ca}[u, v]$. If a spatial domain signal is real-valued, then its frequency-domain counterpart is conjugate symmetric~\cite{dspbook}:

\begin{equation}
    \mathcal{F}\left\{x_\mathfrak{re}[m, n]\right\}=X_\text{cs}[u, v]=\frac{X[u, v]+X^*[u, v]}{2},
    \label{eq:real-cs}
\end{equation}

\noindent where $X^*[u, v]=X_\mathfrak{re}[u, v]-jX_\mathfrak{im}[u, v]$ is the conjugate of $X[u, v]$.

\begin{figure}
    \centering
        \includegraphics[width=\textwidth]{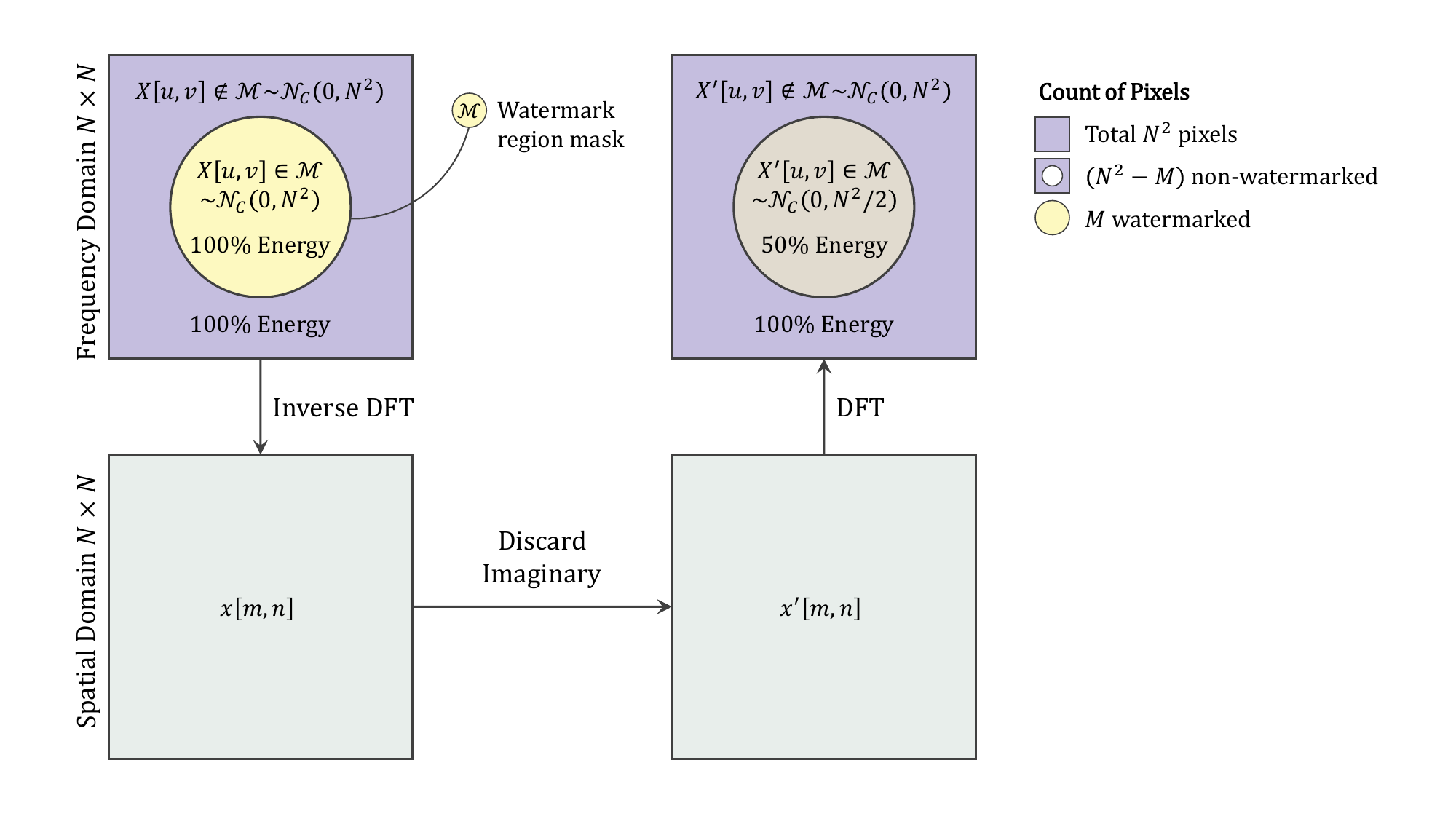}
    \caption{Visualisation of the analysis. {\it Tree-Ring} discards the imaginary part of the spatial domain initial latent noise $x[m, n]$ for diffusion denoising, but this operation discards half of the energy from all watermarked frequency-domain pixels. These pixels also have their variance reduced by half.}
    \label{fig:supp_energy}
\end{figure}

\subsection{Tree-Ring's Pipeline}

As visualized in ~\cref{fig:supp_energy}, {\it Tree-Ring} proposes to add a watermark to a frequency domain initial latent noise $X[u, v]$ (of size $N\times N$) by substituting the value of $M$ pixels within a watermark region mask $\mathcal{M}$. For these watermarked pixels $X[u, v]\in \mathcal{M}$ (corresponds with $w$ used in the paper main content), {\it Tree-Ring} samples their values from a circularly-symmetric complex normal distribution $\mathcal{N}_\mathcal{C}\left(0, N^2\right)=\mathcal{N}\left(0, \frac{N^2}{2}\right)+j\mathcal{N}\left(0, \frac{N^2}{2}\right)$. Other non-watermarked pixels $X[u, v]\notin\mathcal{M}$, individually, also follow this distribution, but the difference is that in {\it Tree-Ring}'s context, $X[u, v]\notin\mathcal{M}$ are spatially ensured to be conjugate symmetric $X[u, v] = X_{cs}[u, v]$, while $X[u, v]\in\mathcal{M}$ does not have such a guarantee. 
During the analysis, we view each pixel as a random variable. 

After adding the watermark to $X[u, v]$, {\it Tree-Ring} transforms it back to the spatial domain, $x[m, n]=\mathcal{F}^{-1}\left\{X[u, v]\right\}$. The obtained $x[m, n]$ would typically have both the real and imaginary parts. However, since diffusion denoising would always start with a purely-real noise signal, {\it Tree-Ring} discards the imaginary part of $x[m, n]$, turning it into $x'[m, n]=x_\mathfrak{re}[m, n]$, whose frequency domain counterpart $X'[u, v]=\mathcal{F}\left\{x'[m, n]\right\}$ is the conjugate symmetric part of the original watermarked signal, $X'[u, v]=X_\text{cs}[u, v]$.

\subsection{Distribution Shift in Watermarked Region}

\cref{eq:real-cs} implies that the frequency domain signal $X'[u, v]$ satisfies:

\begin{equation}
    X'[u, v]=\mathcal{F}\left\{x'[m, n]\right\}=\mathcal{F}\left\{x_\mathfrak{re}[m, n]\right\}=X_\text{cs}[u, v]=\frac{X[u, v]+X^*[-u, -v]}{2}.
\end{equation}

For pixels within the watermark region $X'[u, v]\in\mathcal{M}$, $X[u, v]$ and $X^*[-u, -v]$ are uncorrelated, and they both follow a circularly-symmetric complex normal distribution $\mathcal{N_C}\left(0, N^2\right)$. Viewing $X'[u, v]$ as a random variable, its distribution is given by

\begin{equation}
    X'[u, v]\sim\frac{1}{2}\left(\mathcal{N_C}\left(0, N^2\right)+\mathcal{N_C}\left(0, N^2\right)\right)
    =\mathcal{N_C}\left(0, \left(\frac{1}{2}\right)^2\cdot 2N^2\right)=\mathcal{N_C}\left(0, \frac{N^2}{2}\right).
\end{equation}

Oppositely, pixels outside the watermark region $X'[u, v]\notin\mathcal{M}$ are conjugate symmetric, which implies that $X'[u, v]=X_\text{cs}[u, v]=X[u, v]\sim\mathcal{N_C}\left(0, N^2\right)$. The distribution of non-watermarked pixels $X'[u, v]\notin\mathcal{M}$ are unchanged.




This shows that discarding the imaginary part from the spatial domain pixels $x[m, n]$ changes the distribution of the frequency domain pixels within the watermark region mask $\mathcal{M}$ from $X[u, v]\sim\mathcal{N_C}\left(0, N^2\right)$ to $X'[u, v]\sim\mathcal{N_C}\left(0, \frac{N^2}{2}\right)$.

\subsection{From Energy's Viewpoint}

Discarding the imaginary part of $x[m, n]$ also causes watermarked region $X'[u, v]\in\mathcal{M}$ to lose half of its energy. Since $X[u, v]\sim\mathcal{N_C}\left(0, N^2\right)$, $\lvert X[u, v]\rvert^2\sim N^2\chi^2(1)$. Therefore, the expected energy of $X[u, v]\in\mathcal{M}$ is given by

\begin{equation}
    \mathbb{E}\left[\sum_{u, v\in\mathcal{M}}\lvert X[u, v]\rvert^2\right]
    =(N^2-M)\mathbb{E}_{u, v\in\mathcal{M}}\left[\lvert X[u, v]\rvert^2\right]=(N^2-M)N^2.
\end{equation}

Similarly, $\lvert X'[u, v]\rvert^2\sim \frac{N^2}{2}\chi^2(1)$ for $X'[u, v]\in\mathcal{M}$. Therefore, the expected energy of $X'[u, v]\in\mathcal{M}$ is given by

\begin{equation}
    \mathbb{E}\left[\sum_{u, v\in\mathcal{M}}\lvert X'[u, v]\rvert^2\right]
    =(N^2-M)\mathbb{E}_{u, v\in\mathcal{M}}\left[\lvert X'[u, v]\rvert^2\right]=(N^2-M)\frac{N^2}{2}.
\end{equation}

The fraction of energy of $X'[u, v]\in\mathcal{M}$, compared to $X[u, v]\in\mathcal{M}$, is given by

\begin{equation}
    \eta=\frac{\mathbb{E}\left[\sum_{u, v\in\mathcal{M}}\lvert X'[u, v]\rvert^2\right]}{\mathbb{E}\left[\sum_{u, v\in\mathcal{M}}\lvert X[u, v]\rvert^2\right]}=\frac{(N^2-M)\frac{N^2}{2}}{(N^2-M)N^2}=\frac{1}{2},
\end{equation}

\noindent
So the watermarked region loses half of its energy.

\subsection{Distribution Shift in $\ell_1$ Distance}

In this section, we clarify the systematic $\ell_1$-to-reference shift introduced by discarding the imaginary part during the watermark injection process. To simplify the math, we assume that the recovered watermark comes from the same distribution as the injected one.
We consider four different watermarks $\in\mathcal{M}$ in frequency domain:

\begin{itemize}
    \item $\hat{X}[u,v]\sim \mathcal{N_C}(0, N^2)$: 
    Recovered watermark that never experience imaginary part discarding.
    \item $\hat{X}'[u,v]\sim \mathcal{N_C}(0, \frac{N^2}{2})$:
    Recovered watermark that experienced imaginary part discarding.
    \item $\hat{Y}[u,v]\sim \mathcal{N_C}(0, N^2)$:  Null watermark recovered from unwatermarked images.
    \item $Z[u,v]\sim \mathcal{N_C}(0, N^2)$: The reference watermark to imprint.
\end{itemize}



Since $\hat{X}'_{\mathfrak{re}}[u,v] \sim \mathcal{N}\left(0, \left(\frac{N}{2}\right)^2\right)$ and $Z_{\mathfrak{re}}[u,v] \sim \mathcal{N}\left(0, \left(\frac{N}{\sqrt{2}}\right)^2\right)$ are both Gaussian, their combination is also Gaussian with summed variance:

\begin{equation}
    \left(\hat{X}'_{\mathfrak{re}}[u,v]\pm Z_{\mathfrak{re}}[u,v]\right)\sim\mathcal{N}\left(0, \left(\frac{N}{2}\right)^2+\left(\frac{N}{\sqrt{2}}\right)^2\right)=\mathcal{N}\left(0, \left(\frac{\sqrt{3}}{2}N\right)^2\right).
\end{equation}

And similarly for the imaginary parts. Therefore, 

\begin{equation}
    \left(\hat{X}'_{\mathfrak{re}}[u,v]\pm Z_{\mathfrak{re}}[u,v]\right)^2+\left(\hat{X}'_{\mathfrak{im}}[u,v]\pm Z_{\mathfrak{im}}[u,v]\right)^2\sim\left(\frac{\sqrt{3}}{2}N\right)^2\chi^2(2),
\end{equation}

\noindent where the pdf of a $\chi^2\left(2\right)$-distributed random variable is given by:

\begin{equation}
    f_{\chi^2(2)}(x)=\frac{e^{-\frac{x}{2}}}{2\Gamma(1)}
\end{equation}

Hence, the expected $\ell_1$ distance between $\hat{X}'[u, v]\in\mathcal{M}$ and $Z[u, v]\in\mathcal{M}$, normalised by the count of watermarked pixels $M$, is given by:


\begin{equation}
    \begin{aligned}
        \mathbb{E}\left[\frac{1}{M}\lVert \hat{X}'-Z\rVert_1\right]
        &=\mathbb{E}\left[\frac{1}{M}\sum_{u,v\in\mathcal{M}}\lvert \hat{X}'[u,v]-Z[u,v]\rvert\right]\\
        &=\mathbb{E}\left[\frac{1}{M}\sum_{u,v\in\mathcal{M}}\lvert(\hat{X}'_{\mathfrak{re}}[u,v]+j\hat{X}'_{\mathfrak{im}}[u,v])-(Z_{\mathfrak{re}}[u,v]+jZ_{\mathfrak{im}}[u,v])\rvert\right]\\
        &=\mathbb{E}\left[\frac{1}{M}\sum_{u,v\in\mathcal{M}}\sqrt{\left(\hat{X}'_{\mathfrak{re}}[u,v]-Z_{\mathfrak{re}}[u,v]\right)^2+\left(\hat{X}'_{\mathfrak{im}}[u,v]-Z_{\mathfrak{im}}[u,v]\right)^2}\right]\\
        &=\mathbb{E}_{u,v\in\mathcal{M}}\left[\sqrt{\left(\hat{X}'_{\mathfrak{re}}[u,v]-Z_{\mathfrak{re}}[u,v]\right)^2+\left(\hat{X}'_{\mathfrak{im}}[u,v]-Z_{\mathfrak{im}}[u,v]\right)^2}\right]\\
        &=\frac{\sqrt{3}}{2}N\int_\mathbb{R} \sqrt{x}f_{\chi^2(2)}(x)dx=\frac{\sqrt{3}}{2}N\sqrt{\frac{\pi}{2}}.
    \end{aligned}
\end{equation}

Similarly, making use of $\hat{X}_{\mathfrak{re}}[u,v], \hat{X}_{\mathfrak{im}}[u,v], \hat{Y}_{\mathfrak{re}}[u,v], \hat{Y}_{\mathfrak{im}}[u,v], Z_{\mathfrak{re}}[u,v], Z_{\mathfrak{im}}[u,v]\sim \mathcal{N}\left(0, \left(\frac{N}{\sqrt{2}}\right)^2\right)$, the pixel-number-normalized $\ell_1$ distance is given by

\begin{equation}
    \mathbb{E}\left[\frac{1}{N^2}\lVert \hat{X}[u,v]-Z[u,v]\rVert_1\right]=N\sqrt{\frac{\pi}{2}}.
\end{equation}

\begin{equation}
    \mathbb{E}\left[\frac{1}{N^2}\lVert \hat{Y}[u,v]-Z[u,v]\rVert_1\right]=N\sqrt{\frac{\pi}{2}}.
\end{equation}

These expectations satisfy:

\begin{equation}
\label{eq:supp_shift_conclusion}
    \mathbb{E}\left[\lVert \hat{X}'-Z\rVert_1\right]=\frac{\sqrt{3}}{2}\mathbb{E}\left[\lVert \hat{X}-Z\rVert_1\right]=\frac{\sqrt{3}}{2}\mathbb{E}\left[\lVert \hat{Y}-Z\rVert_1\right].
\end{equation}

\textbf{Generally speaking, discarding the imaginary part causes the expectation of $\ell_1$-to-reference distance to shift by $\frac{\sqrt{3}}{2}$ statistically.} This shift factor is derived under several assumptions. In practice, situations are more complicated. \textit{Tree-Ring} injects a fixed watermark pattern. 
Thus $\hat{X}$, $\hat{X}'$ and $Z$ all carries information of a specific pattern and correlates with each other. They are not i.i.d samples.
Both pattern matching and the distribution shift contributes to the expectation of $\ell_1$ distance in~\cref{eq:supp_shift_conclusion}. 




\subsection{Distribution Shift in Real Scenarios}
\label{subsec:supp_shift_in_real}
As mentioned above, in practice, many other factors affect the distribution shift of the $\ell_1$ distance to reference. These factors include the matching of pattern, attacks, diffusion and inversion process, etc. To assess the distribution shift in real scenarios, we conduct a set of control experiments.

In Control 1, we follow the original setup of {\it Tree-Ring}. The original setup aims at distinguishing between the recovered watermark $\hat{w}$ (shifted) and the null watermark $\hat{w}_\varnothing$ (not shifted) recovered from the unwatermarked images, \ie $\lVert \hat{w} -w \rVert_1$ {\it v.s.} $\lVert \hat{w}_\varnothing -w \rVert_1$.

In Control 2, we shift null watermark $\hat{w}_\varnothing$ to the same extent as the operation of discarding imaginary part does. Then we get a shifted null watermark $\hat{w}_\varnothing'$. We distinguish between $\hat{w}$ and $\hat{w}_\varnothing'$,  \ie $\lVert \hat{w} -w \rVert_1$ {\it v.s.} $\lVert \hat{w}_\varnothing' -w \rVert_1$. Here, both $\hat{w}$ and $\hat{w}_\varnothing'$ are shifted to the same extent, so we eliminate the help of distribution shift. 

We compare the results of Control 1 and 2 in~\cref{tab:supp_control_auc}. We can find general performance drop under all attacks in Control 2. The average AUC decreases from 0.975 to 0.913, making it more challenging to distinguish watermarked and non-watermarked images without the help of distribution shift. Further observation reveals that the major drop occurs in Rotation and Crop \& Scale attacks, indicating that distribution shift contributes substantially to the robustness to Rotate and Crop \& Scale attacks. This also implies that the original {\it Tree-Ring} watermark pattern cannot handle these attacks.




Meanwhile, we stat the average $\ell_1$-to-reference distance in these two control experiments and compare them in~\cref{tab:supp_control_l1}. Without the help of distribution shift in Control 2, the expectations of $\lVert \hat{w}_\varnothing -w \rVert_1$ and $\lVert \hat{w}_\varnothing' -w \rVert_1$ are closer, especially under Rotation and C\&S attacks, indicating overlapped distribution. This implies a lower AUC and more difficulty in distinguishing, consistent with the results in~\cref{tab:supp_control_auc}.


\begin{table}[tb]
  \caption{Control experiments to demonstrate the effect of distribution shift. We report AUC in verification setting. In Control 1, distribution shift helps distinguish. In Control 2, distribution shift doesn't help. We observe the big performance drop under Rotate and C\&S attacks.}
  \label{tab:supp_control_auc}
  \centering
  \setlength{\tabcolsep}{4.5pt}
  \resizebox{\linewidth}{!}{
  \begin{tabular}{c|ccccccc|c} 
    \toprule[1.1pt]
    \textbf{Experiment} & \textbf{Clean} & \textbf{Rotate} & \textbf{JPEG} & \textbf{C\&S} & \textbf{Blur} & \textbf{Noise} & \textbf{Bright} & \textbf{Avg}\\
    \midrule
    Control 1    &\textbf{1.000} &\textbf{0.935} &\textbf{0.999} &\textbf{0.961} &\textbf{0.999} & \textbf{0.944} & \textbf{0.983} &\textbf{0.975} \\
    Control 2   &\textbf{1.000 }&\textcolor{blue}{0.728} &\textbf{0.999} &\textcolor{blue}{0.746} &0.998 &0.940 &0.978 & 0.913 \\
  \bottomrule[1.1pt]
  \end{tabular}
  }
\end{table}

\begin{table}[tb]
  \caption{Average $\ell_1$-to-reference distance in 2 control experiments. We use $\Delta$ to denote the difference between $\lVert \hat{w}_\varnothing -w \rVert_1$ and $\lVert \hat{w}_\varnothing -w \rVert_1$, $\lVert \hat{w}_\varnothing -w \rVert_1$ and $\lVert \hat{w}_\varnothing' -w \rVert_1$. Larger $\Delta$ means easier to distinguish. }
  \label{tab:supp_control_l1}
  \centering
  \setlength{\tabcolsep}{4pt}
  \resizebox{\linewidth}{!}{
  \begin{tabular}{c|c|ccccccc|c} 
    \toprule[1.1pt]
    \textbf{Experiment} & \textbf{Target} & \textbf{Clean} & \textbf{Rotate} & \textbf{JPEG} & \textbf{C\&S} & \textbf{Blur} & \textbf{Noise} & \textbf{Bright} & \textbf{Avg}\\
    \midrule
    - &  $\lVert \hat{w} -w \rVert_1$ & 51.50 & 78.85 & 66.16 & 76.96 & 69.41 & 73.73 & 64.86 & 68.78  \\
    \midrule
    \multirow{2}{*}{{Control 1}}
        & $\lVert \hat{w}_\varnothing -w \rVert_1$  & 83.92 & 83.96 & 84.35 & 84.58 & 90.84 & 83.10 & 83.30 & 84.86 \\
        & $\Delta$   & 32.43 & 5.11 & 18.18 & 7.63 & 21.43& 9.37 & 18.44 & \textbf{16.08 }\\
    \midrule
    \multirow{2}{*}{{Control 2}}
         & $\lVert \hat{w}_\varnothing' -w \rVert_1$   & 77.31 & 80.60 & 81.21 & 79.70 & 86.46 & 81.63 & 80.34 & 81.04 \\
         & $\Delta$   & 25.81 & \textcolor{blue}{1.75} & 15.05 & \textcolor{blue}{2.74} & 17.05 & 7.90 & 15.48 & \underline{12.26} \\
  \bottomrule[1.1pt]
  \end{tabular}
  }
\end{table}

\section{Discarding Imaginary Part As Standalone Watermarking Approach}
\label{sec:supp_standalone_watermark}
We demonstrate the operation of discarding the imaginary part can be used as a standalone watermarking approach.
Concretely, for each initial noise instance intended for watermarking, rather than injecting a ring watermark into the frequency spectrum $X$, we opt to inject a random Gaussian noise $\sim \mathcal{N_C}(0, N)$ into the same region. Note that the injected noise are {\it i.i.d.} sampled for each case thus different from each other. Although originating from the same distribution, the newly introduced Gaussian noise typically lacks conjugate symmetry. Consequently, when transforming to the spatial domain, it necessitates discarding the excess imaginary parts. This results in a distribution shift in watermarked noise, thus the actually injected noise is from $\mathcal{N_C}(0, \frac{N}{2})$.

As discussed in previous sections, this shift induces deviations in the $\ell_1$ distance and energy of the watermarked noise from the non-watermarked noise. During verification, we explore three different methods to distinguish the watermarked and the non-watermarked. Specifically, we compute the $\ell_1$ distance between the recovered noise and three different references: (1) random Gaussian noise $\sim \mathcal{N_C}(0, N)$ (2) zero (3) zero but with $\ell_2$ distance (corresponding to distinguishing the energy between the watermarked and the non-watermarked). Empirical results are presented in~\cref{tab:supp_standalone}.
We can find all the mentioned methods effectively detect the presence of the watermark. When we use zero as the reference, the distinction is most pronounced, highlighting its superior discriminatory performance.

\noindent
\textbf{Relation with \textit{\texorpdfstring{Tree-Ring\textsubscript{rand}}{Tree-Ring Rand}}} \cite{wen2023tree} provides a variant called \textit{\texorpdfstring{Tree-Ring\textsubscript{rand}}{Tree-Ring Rand}} that also injects noise as the watermark. However, they inject the same noise pattern for all generated images and intend to rely on pattern matching for watermark verification. The proposed method in this section distinguishes itself from \textit{\texorpdfstring{Tree-Ring\textsubscript{rand}}{Tree-Ring Rand}} by injecting {\it i.i.d.} sampled noise for each generated image. The AUC for \textit{\texorpdfstring{Tree-Ring\textsubscript{rand}}{Tree-Ring Rand}}~\cite{wen2023tree} and ours is 0.918 and 0.901, respectively. The closely matched performances indicate that the deviation introduced by discarding imaginary part offers very robust discriminative power. 
This suggests that discarding imaginary part can effectively distinguish between watermarked and non-watermarked images even without relying on the specific noise pattern.

\begin{table}[tb]
  \caption{Quantitative results when discarding imaginary part is used as a standalone watermarking approach. We calculate the distance between the recovered noise and different references for distinguishment between the watermarked and the non-watermarked. Note that when the reference is Zero and the metric is $\ell_2$, it actually distinguishes by energy. AUC is reported.}
  \label{tab:supp_standalone}
  \centering
  \setlength{\tabcolsep}{3.5pt}
  \resizebox{\linewidth}{!}{
  \begin{tabular}{c|c|ccccccc|c} 
    \toprule[1.1pt]
    \textbf{Ref} & \textbf{Metric} & \textbf{Clean} & \textbf{Rotate} & \textbf{JPEG} & \textbf{C\&S} & \textbf{Blur} & \textbf{Noise} & \textbf{Bright} & \textbf{Avg}\\
    \midrule
    Gaussian & $\ell_1$       &0.970 &0.837 &0.812 &0.907 &0.841 &0.663 &0.783 &0.831 \\
    Zero &  $\ell_1$     &0.998 &\textbf{0.908} &\textbf{0.925} &\textbf{0.967} &0.879 &\textbf{0.747} &\textbf{0.881} & \textbf{0.901} \\
    Zero &  $\ell_2$       &\textbf{0.999} & \textbf{0.908} &\textbf{0.925} &\textbf{0.967} &\textbf{0.883} &0.741 &\textbf{0.881} & \textbf{0.901} \\
  \bottomrule[1.1pt]
  \end{tabular}
  }
\end{table}

\section{Failure Cases of Multi-Channel Rings}
\label{sec:supp_failure_case}
\begin{figure}[tb]
  \centering
  \includegraphics[width=\linewidth]{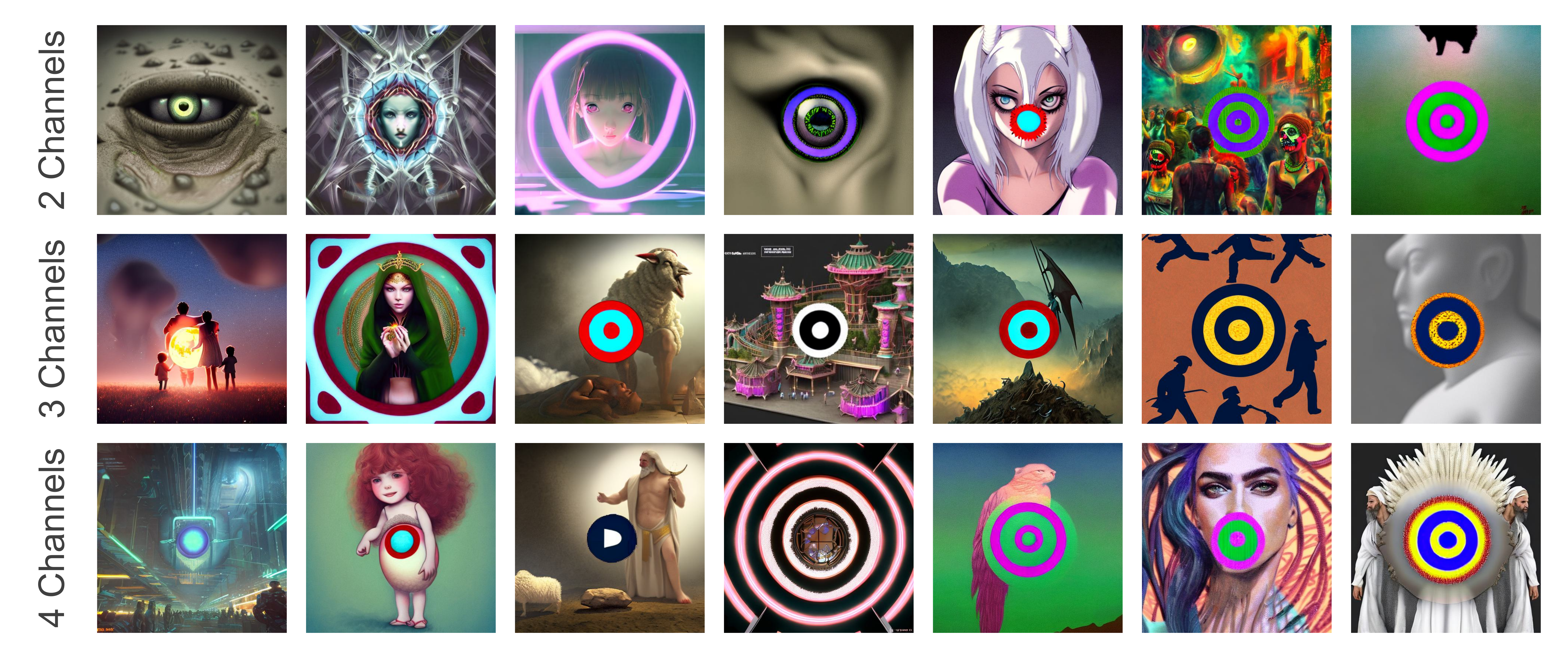}
  \caption{Generated artifacts when we imprint the ring watermarks on multiple channels.}
  \label{fig:failure_case}
\end{figure}

As discussed in the main text, we can imprint the ring watermarks onto multiple channels to increase the capacity. However, we find that this often leads to the generation of ring-like artifacts, evident in a substantial proportion of cases, illustrated in \cref{fig:failure_case}. So we only imprint the ring watermark on a single channel by default.

\begin{table}[tb]
    \caption{Empirical results of {\it RingID} on different diffusion models. Identification accuracy is reported.}
  \vspace{-0.3cm}
  \label{tab:supp_ablation_sd}
  \centering
  \setlength{\tabcolsep}{3pt}
  \resizebox{\linewidth}{!}{%
  \begin{tabular}{c|c|ccccccc|c} 
    \toprule[1.1pt]
    \textbf{Models} & \textbf{\#Assigned Keys} & \textbf{Clean} & \textbf{Rotate} & \textbf{JPEG} & \textbf{C\&S} & \textbf{Blur} & \textbf{Noise} & \textbf{Bright} & \textbf{Avg}\\
    \midrule
    SD 1.4      &128              &0.950 &0.920 &0.950 & \textbf{0.350} &0.950 &0.930 &0.910 &0.851 \\
    SD 1.5      &128              &0.960 &0.940 &0.960 &0.340 &0.950 &0.940 &0.910 &0.857 \\
    SD 2.1      & 128 & \textbf{1.000} & \textbf{0.980} & \textbf{1.000} & 0.280 & \textbf{0.980} & \textbf{1.000} & \textbf{0.940} & \textbf{0.883} \\
    \midrule
    SD 1.4      &2048             &0.970 &0.810 &0.950 &0.080 &0.970 &0.900 &0.820 &0.786 \\
    SD 1.5      &2048             &0.990 &0.800 &0.970 &\textbf{0.110} &0.950 &\textbf{0.950} &0.850 &0.803 \\
    SD 2.1      & 2048 & \textbf{1.000} & \textbf{0.860} & \textbf{1.000} & 0.080 & \textbf{0.970} & \textbf{0.950} & \textbf{0.870} & \textbf{0.819} \\
  \bottomrule[1.1pt]
  \end{tabular}}
\end{table}

\section{More Ablations and Comparisons}
\subsection{Results on Different Diffusion Models}
{\it RingID} is a universal method that can be applied to different diffusion models. \cref{tab:supp_ablation_sd} shows the results on more diffusion models. It is worth noting that the performance of {\it RingID} gradually improves from the older version of SD to the newer version of SD.

\subsection{Comparison with More Methods}
\cref{tab:supp_more_methods_verification} compares {\it \ours} with more methods on the watermark verification task. We can find that {\it \ours} achieves the best \textbf{AUC} and \textbf{TPR@1\%FPR} under both clean and adversarial settings, showcasing strong robustness.

\begin{table}[t]
    \centering
        \caption{Comparison with more methods on verification. }
        \label{tab:supp_more_methods_verification}
        \setlength{\tabcolsep}{2pt}
        \resizebox{\linewidth}{!}{%
        \begin{tabular}{c|c|c|c|c}
        \toprule[1.1pt]
        \textbf{Methods} & \textbf{AUC/T@1\%F (Clean)}$\uparrow$ & \textbf{AUC/T@1\%F (Adv)}$\uparrow$ & \textbf{FID$\downarrow$} & \textbf{CLIP Score $\uparrow$} \\
        \midrule[0.75pt]
        {\it DwtDctSvd}~\cite{cox2007digital} & \textbf{1.0} / \textbf{1.0} &  0.702 / 0.262 & 25.01 & 0.359 \\
        {\it RivaGAN}~\cite{zhang2019robust} & 0.999 / 0.999 & 0.854 / 0.448 & \textbf{24.51} & 0.361 \\
        {\it Tree-Ring}~\cite{wen2023tree} & \textbf{1.0} / \textbf{1.0} & 0.975 / 0.694 & 25.93 & 0.364 \\
        \rowcolor{mygray}
        {\it RingID} & \textbf{1.0} / \textbf{1.0} & \textbf{0.995} / \textbf{0.926} & 26.13 & \textbf{0.365} \\
        \bottomrule[1.1pt]
        \end{tabular}}
    \vspace{-0.3cm}
\end{table}

\section{More Qualitatives}
\begin{figure}[tb]
  \centering
  \includegraphics[width=\linewidth]{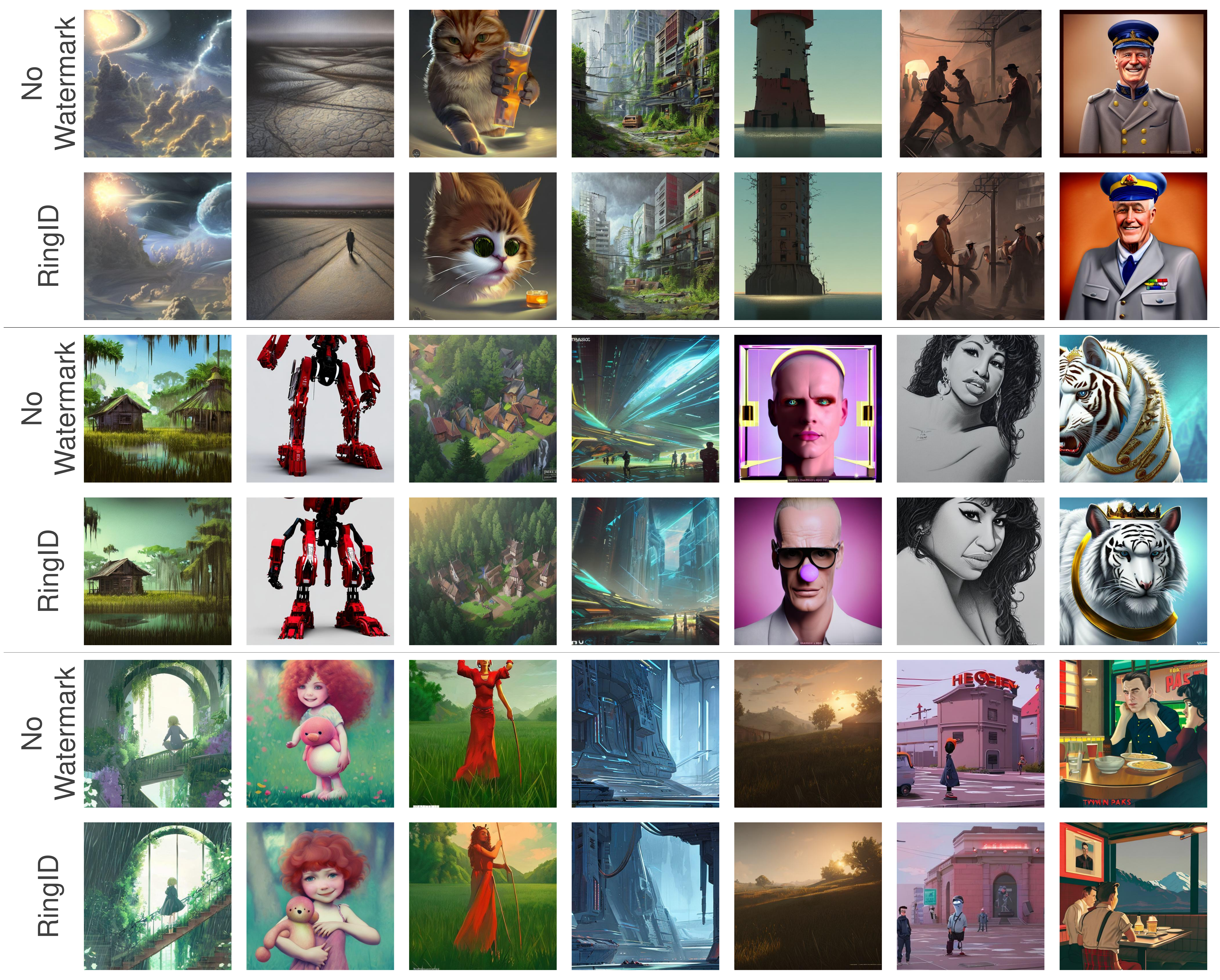}
  \caption{More qualitative results. Images are generated by SD 2.1 with Stable-Diffusion-Prompts~\cite{sdprompts}.}
  \label{fig:supp_more_qualitatives}
\end{figure}

\cref{fig:supp_more_qualitatives} gives more qualitative results.

\end{document}